# DANet: Enhancing Small Object Detection through an Efficient Deformable Attention Network

Md Sohag Mia†
School of Artificial Intelligence
Nanjing University of
Information Science and
Technology
Nanjing, China
shuvosohagahmmed@gmail.com

Abdullah Al Bary Voban *
School of Artificial Intelligence
Nanjing University of
Information Science and
Technology
Nanjing, China
voban@nuist.edu.cn

Abu Bakor Hayat Arnob *
School of Artificial Intelligence
Nanjing University of
Information Science and
Technology
Nanjing, China
satcarnob@gmail.com

Abdu Naim*
School of Artificial Intelligence
Nanjing University of
Information Science and
Technology
Nanjing, China
naimabdu@nuist.edu.cn

Md Kawsar Ahmed
School of Computer Science
Beijing Institute of Technology
Beijing, China
mkawsarcse@gmail.com

Md Shariful Islam
School of Computer Science
Beijing Institute of Technology
Beijing, China
sharif@bit.edu.cn

*Abstract*—Efficient and accurate detection of small objects in manufacturing settings, such as defects and cracks, is crucial for ensuring product quality and safety. To address this issue, we proposed a comprehensive strategy by synergizing Faster R-CNN with cutting-edge methods. By combining Faster R-CNN with Feature Pyramid Network, we enable the model to efficiently handle multi-scale features intrinsic to manufacturing environments. Additionally, Deformable Net is used that contorts and conforms to the geometric variations of defects, bringing precision in detecting even the minuscule and complex features. Then, we incorporated an attention mechanism called Convolutional Block Attention Module in each block of our base ResNet50 network to selectively emphasize informative features and suppress less useful ones. After that we incorporated RoI Align, replacing RoI Pooling for finer region-of-interest alignment and finally the integration of Focal Loss effectively handles class imbalance, crucial for rare defect occurrences. The rigorous evaluation of our model on both the NEU-DET and Pascal VOC datasets underscores its robust performance and generalization capabilities. On the NEU-DET dataset, our model exhibited a profound understanding of steel defects, achieving state-of-the-art accuracy in identifying various defects. Simultaneously, when evaluated on the Pascal VOC dataset, our model showcases its ability to detect objects across a wide spectrum of categories within complex and small scenes.

*Keywords— Faster R-CNN, DCN, CBAM, Small Object Detection.*

## I. Introduction

In the realm of modern manufacturing, the pursuit of enhanced quality control and production efficiency has been fuelled by the integration of cutting-edge technologies, most notably deep learning. This paper delves into the domain of small object detection within manufacturing factories, focusing specifically on small objects, defects, and crack detection. The ability to detect imperfections at a microscopic scale holds paramount significance in ensuring product reliability and safety. Defect detection methods have recently shifted from manual procedures [1] to ones based on computer vision. The object detection strategy for detecting surface defects through computer vision predominantly relies on features. Extracting these features involves the use of algorithms that are manually constructed, leading to weaknesses in the model's robustness and its ability to generalize. Deep learning methods come to the rescue in such situations. Convolutional neural networks can capture high-level semantic feature maps of input images, leading to models that possess better robustness and generalization capabilities than traditional methods. A thorough analysis of the existing works reveals that while numerous approaches have been proposed, a significant gap remains between their capabilities and the demands of a real-time, high-accuracy defect detection system. This void serves as the impetus for our current study, wherein we introduce a novel model named DANet or **D**eformable **A**ttention **Net**work to address these limitations comprehensively. In the subsequent sections, we will delve into a comprehensive discussion of the prevalent approaches within the field of steel surface defect detection. By examining their strengths and shortcomings, we will underscore the pressing need for an innovation that combines both real-time processing and exceptional accuracy. Our proposed DANet model emerges as a solution to bridge this gap, offering significant contributions that promise to redefine the landscape of steel surface defect detection. The main contributions of our work are as follows. At first, we innovatively integrated the Deformable Convolutional Network (DCN) [2] into the ResNet50 [3] model, effectively replacing the conventional fixed convolutional layers with adaptable deformable layers. Building upon this, we advanced the model by incorporating the Convolutional Block Attention Module (CBAM) [4] into the DCN-ResNet architecture. This strategic addition enhances the generation of more intricate and semantically meaningful feature maps. To further amplify performance, we employed the Feature Pyramid Network (FPN) [5] to seamlessly fuse multiscale features, allowing our model to robustly capture features across different levels of abstraction. Addressing the challenge of class imbalance, we strategically integrated the Focal Loss module [6]. This addition significantly improves the model's ability to handle uneven class distributions and focus more on challenging instances. In summary, our contributions encompass a holistic approach that encompasses the integration of DCN, CBAM, FPN, and Focal Loss into the Faster R-CNN framework, resulting in a more powerful and effective model for tackling complex defect detection tasks.

† Leading contributor.   ∗ Equal contribution.

## II. RELATED WORKS

The loss function for the training of the detection model is presented as an optimized IOU. In order to increase detection accuracy, Xiao et al. [7] and Jie [8] suggest a surface defect detection method for hot-rolled steel strip that combines attention mechanism and multi-feature fusion network. In order to filter and retain essential information, the technique incorporates a channel attention mechanism, which speeds up detection and decreases network processing. Low-level and high-level features are merged to complement one another using ResNet-50 as the feature extraction network, increasing the precision of detection. Ouyang et al. [9] proposes an end-to-end defect detection network that can locate the defect position of steel plate images and predict the defect category by introducing a novel feature pyramid network module called the adaptive spatial attention feature pyramid, which effectively fuses texture features at low levels with semantic features at high levels. Furthermore, an adaptive convolution and anchor (ACA) module, which consists of adaptive convolution and adaptive anchor is introduced to improve the performance of metallic surface defect detection. Zhao et al. [10] proposes an improved target detection algorithm for steel surface defect detection, addressing the problem of detecting small and complex defect targets, by introducing several improvements to the traditional Faster R-CNN algorithm. A defect classification method usually includes two parts: a feature extractor and a classifier. The most famous region-based detectors are the "R-CNN family" [11], [12] In this framework, thousands of class-independent region proposals are employed for detection. Region-based methods are superior in precision but require slightly more computation. The representative direct regression methods are YOLO [13] and SSD [14]. They directly divide an image into small grids and then for each grid predict bounding boxes, which then regressed to the ground truth boxes. The direct regression method is fast to detect but struggles in small instances.

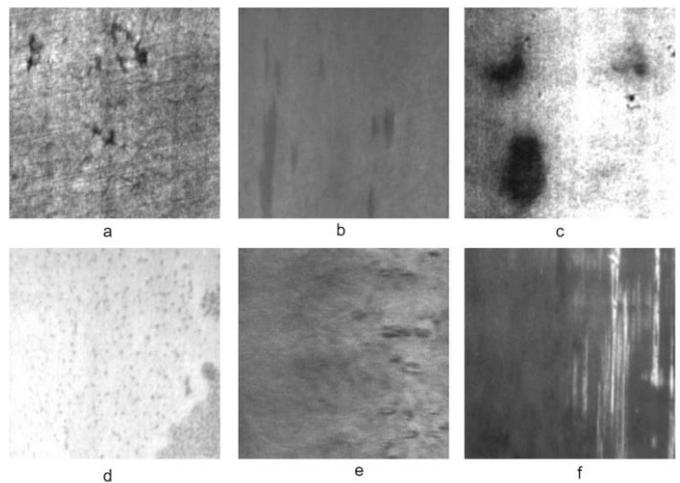

Fig. 1. Sample 6 types of images of NEU-DET dataset. (a) Crazing, (b) Inclusion, (c) Patches, (d) Pitted_surface, (e) Rolled_in_scale, (f) Scratches.

### A. Choice of FRCNN Over YOLO Series Methods

Small object detection in manufacturing environments requires a tailored approach to handle the unique challenges posed by the detection of minuscule and complex defects. While YOLO (You Only Look Once) series methods have gained prominence in the computer vision community for their real-time capabilities, we opted for Faster R-CNN due to several compelling reasons specific to our problem domain.

*1) Model Suitability:* Faster R-CNN's region-based approach aligns well with the intricacies of small and complex defect detection task. In manufacturing environments, small defects often exhibit irregular shapes and may appear at various scales. The region proposal mechanism of Faster R-CNN (FRCNN) allows the model to focus on potential defect regions, making it particularly effective in handling such scenarios.

*2) Comparison:* While direct experimentation comparing FRCNN with YOLO series methods was not the primary focus of this study, we conducted thorough experimentation and benchmarking against a range of object detection models. Our proposed method outperformed these[15],[16],[17] enhanced YOLO series models, yielding superior results. It's showing that our results consistently demonstrated the effectiveness of Faster R-CNN for small object detection in manufacturing environments, which further reinforced our choice.

*3) Better Localization:* YOLO models may struggle with precise object localization, especially for small objects, because they use a grid-based approach. In contrast, Faster R-CNN's

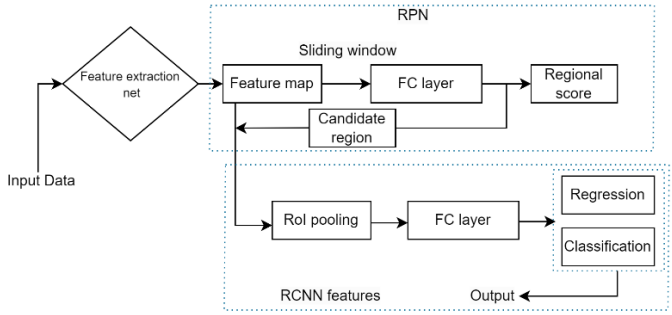

Fig. 2. Flowchart of base Faster R-CNN model.

two-stage architecture is designed to produce more accurate bounding box predictions.

*B. Transformer Based Methods*

In recent years, Transformer-based detection networks, such as DETR[18], have garnered significant attention in the computer vision community but it's still not very effective for small object detection[19]. These models have demonstrated exceptional capabilities in various tasks, prompting a natural question regarding their suitability for small object detection in manufacturing environments. To provide insights into this comparison, we conducted a comprehensive evaluation of our Faster R-CNN-based approach in comparison section against Transformer-based alternatives.

In summary, improved CNN methods excel over original Vision Transformers (ViTs) in small object detection with limited data due to their superior performance, generalization capabilities, reduced computational overhead, noise robustness, resource-efficient training, and the support of an existing ecosystem, making it particularly well-suited for real-time applications where rapid and precise small object detection is essential.

### III. PROPOSED FASTER R-CNN MODEL

*A. Architecture of Improved Faster R-CNN*

The Faster R-CNN (Region-based Convolutional Neural Network) [20] architecture revolutionizes object detection by integrating object proposal generation and classification into an end-to-end pipeline. Starting with an input image, a shared convolutional backbone extracts feature maps. These maps are utilized by the Region Proposal Network (RPN) to propose candidate object regions, which are refined through RoI Pooling. The subsequent object classifier performs category prediction and precise bounding box regression. During training, the RPN and classifier are optimized using region-based losses. At inference, the model efficiently processes images, produces accurate object proposals, and conducts classification and localization, making Faster R-CNN a foundational advancement in real-time object detection. The architecture of our base Faster R-CNN model is shown in Fig. 2, and in Fig. 9 we depicted our proposed DANet model.

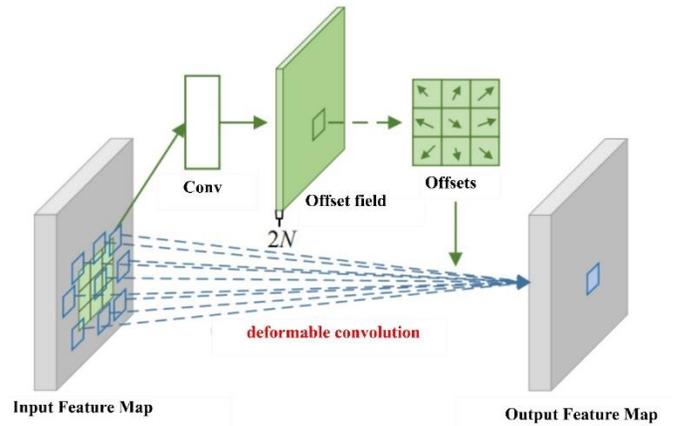

Fig. 3. Architecture of deformable convolutional network.

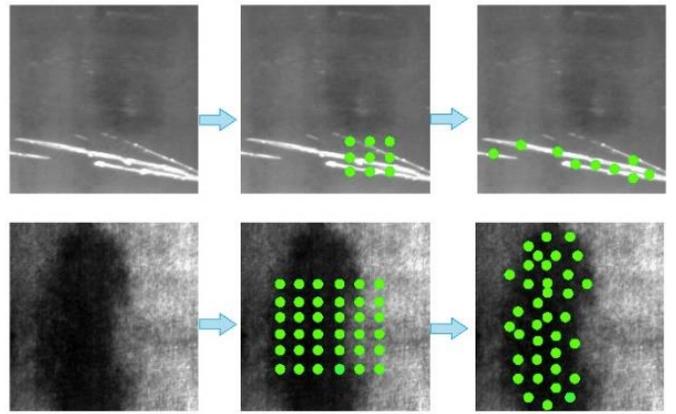

Fig. 4. The calculation system of traditional and deformable convolution.

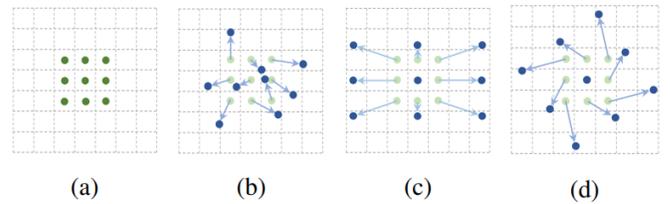

Fig. 5. Four sampling methods of convolution.

*B. Residual Network*

ResNet-50 [3] is a variant of the ResNet architecture that allows for the training of neural networks with hundreds of layers without suffering from the vanishing gradient problem. ResNet-50 introduces the concept of "skip connections" (also called "residual connections"), which allow the network to learn only the difference (residual) between the desired output and the current estimation. ResNet-50's innovation of residual connections allowed it to create deeper neural networks that achieved remarkable performance on object detection benchmarks.

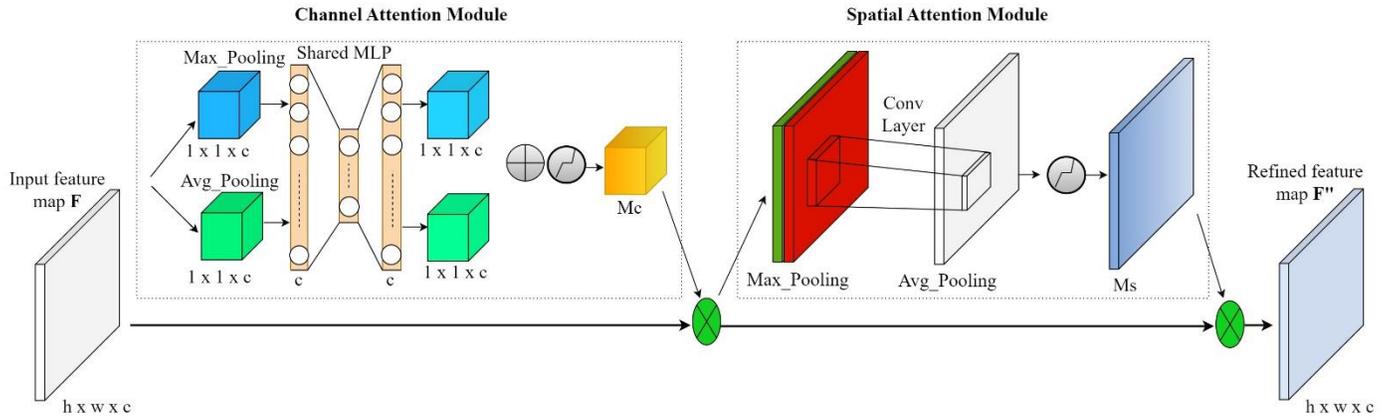

Fig. 6. The structure of CBAM module.

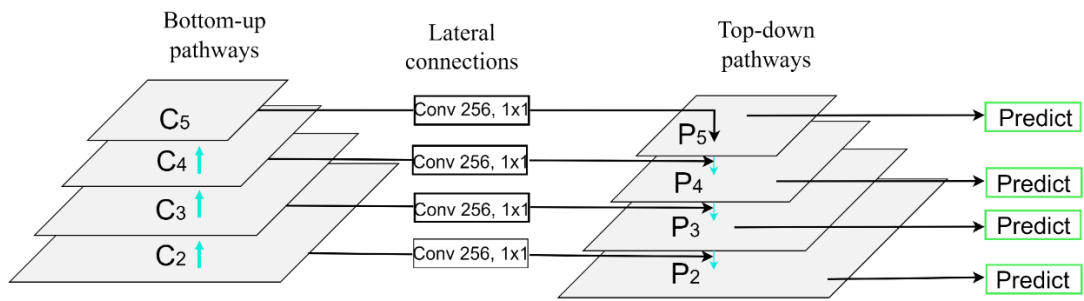

Fig. 7. The Architecture of FPN.

*C. Feature Pyramid Network*

The Feature Pyramid Network (FPN) has emerged as a pivotal architecture in the realm of computer vision, particularly in the field of object detection and segmentation. FPN addresses the challenge of feature representation across multiple scales. The significance of FPN lies in its ability to fuse multi-scale information into a single framework, leading to enhanced detection performance across objects of varying sizes. The core idea of FPN involves creating a feature pyramid from a base convolutional neural network (CNN). This pyramid facilitates the generation of rich feature maps at different scales, catering to the demands of detecting both small and large objects. The FPN architecture is constructed through a top-down and lateral connection approach, which enables the integration of high-resolution features from shallow layers with low- low-resolution, semantically rich features from deeper layers. This process generates a set of feature maps that capture both fine-grained details and contextual information. At each level of the FPN, feature maps are refined using lateral connections. These connections involve convolutions that transform features from shallower layers to align with the scale of features in deeper layers. Subsequently, these refined features are merged with the downsampled features from the previous level, resulting in a fused representation that preserves spatial information while incorporating contextual understanding. This stepwise refinement creates a feature hierarchy, where higher levels of the pyramid capture broader contextual semantics, while lower levels emphasize finer details as depicted in Fig 7. In particular, the proposed backbone network analyses images to create various feature maps of varied resolutions, denoted as $C_i (1 \leq i \geq 5)$ and corresponding to strides of {2,4,8,16,32} pixels with respect to the input image. However, because of its excessive memory usage, $C_1$ was omitted from the feature pyramid network. To obtain $P_5$, FPN first decreases the dimensionality using $1 \times 1$ convolution. It then uses bilinear interpolation to upsample $P_5$ to the same size as $C_4$. For all output channels, the same 256 dimension (number of channels) has been defined. Following that, FPN used $1 \times 1$ convolution to reduce the dimensionality of $P_4$. $P_5$ and $P_4$ didn't undergo any change in size and ended up being identical. By directly incorporating the $P_5$ element into $P_4$, FPN achieved the updated $P_4$. The same technique was subsequently applied to renew $P_3$, and $P_2$. FPN efficiently updated the complete network, following a top-down approach.

*D. Deformable Convolutional Network*

Deep convolutional neural networks have an advantage over traditional methods because they can learn effective features from large amounts of data, without relying on human-designed features. This is made possible through their powerful modeling ability and automatic end-to-end learning approach. Despite their prowess, Convolutional Neural Networks still face a constricting challenge when it comes to modeling geometric transformations, as their building blocks are rigidly fixed, hindering their flexibility to navigate the diverse spatial landscape. To solve the above issue *Deformable ConvNet* is integrated with residual layers for enhancing the transformation modeling capability of CNNs. This approach involves adding offsets to spatial sampling locations and learning them from target tasks, which can be learned without extra supervision. The shape of the convolution kernel in deformable convolution is shown in Figure 5. Figure 5(a) is the standard convolution kernel with the size of $3 \times 3$ (the green dot in the figure); Figure 5b is the deformable convolution. Figures 5(c) and 5(d) are the special cases of deformable convolutional kernel sampling.

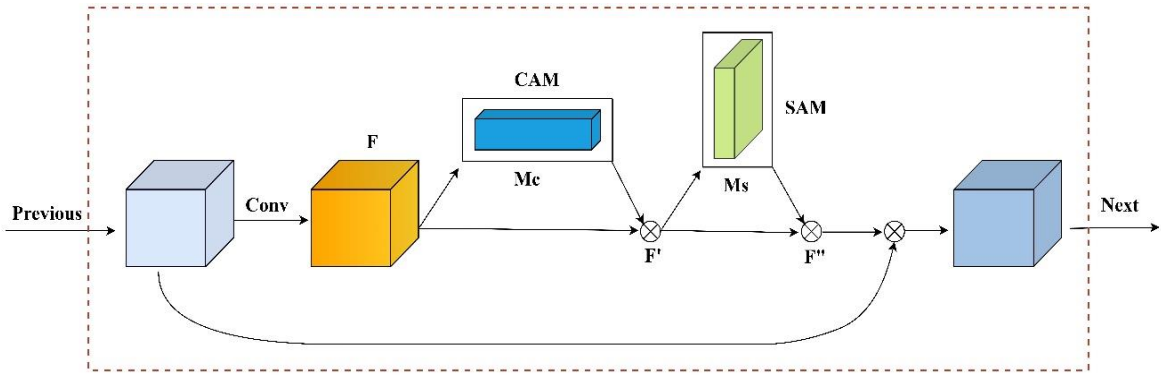

Fig. 8. Architecture of ResNet block with attention modules.

$$y(p_0) = \sum_{p_n \in R} w(p_n) \cdot x(p_0 + p_n), \quad (1)$$

Here, $p_n$ computes the locations within $R$.

However, in deformable convolution, the standard grid $R$ is complemented with offsets $\{\Delta p_n | n = 1, \ldots, N\}$, where $N = |R|$.
The deformation convolution of each pixel in the input image is expressed as follows:

$$y(p_0) = \sum_{p_n \in R} w(p_n) \cdot x(p_0 + p_n + \Delta p_n) \quad (2)$$

In equations (1) and (2), for each location on the output feature map $y$, $x$ is the input feature map and a corresponding location $p_0$ is identified, $w$ is the parameter of weight, $p_n$ is any pixel in convolution, and $\Delta p_n$ is the offset value. The bias domain of deformable convolution pertains to selecting the sampling point with a distinct intention and generating a greater amount of feature information. This is achieved through the manipulation of the sampling position via the bias matrix in convolution, allowing for a more flexible transformation. Consequently, when employing stacked deformable convolutions, there is a significant enhancement in feature extraction capability. To illustrate this point using a 3 × 3 convolution as a case in point, the author employs Figure 5 to show the contrast between standard convolution and deformable convolution. In traditional convolution depicted in Figure 5(1st raw), the sampling position on the target remains fixed. In contrast, deformable convolution depicted in Figure 5(2nd raw) dynamically adapts its receptive field during computation, bestowing it with a robust capacity for extracting features from intricate and irregular targets. This adaptability allows it to conform to the shape and dimensions of the target, thereby underscoring its crucial role in the context of steel defect detection.

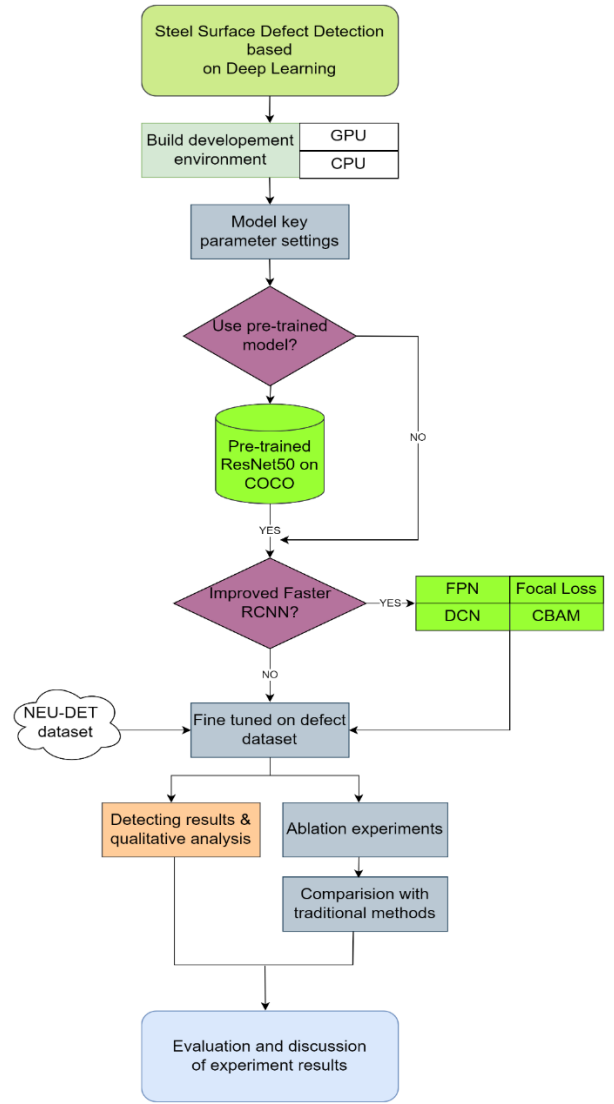

Fig. 9. The Flowchart of our proposed DANet.

*E. Convolutional Block Attention Mechanism*

When dealing with intricate features of input images, it is crucial to enhance the feature extraction performance of the underlying convolutional neural network. We mainly focus on depth, width, and cardinality—the three key elements of networks to improve the performance of CNNs [21][22]. However, there's a hidden gem waiting to be discovered that can significantly boost the efficiency of these CNN models - Attention mechanism. This concept draws inspiration from the realm of cognitive neuroscience. The attention mechanism is a type of processing that may choose and independently learn to pay attention to key aspects. We combined our trusty base ResNet with the CBAM attention model to reap these benefits. The Convolution Block Attention Module (CBAM) is a master of duality, expertly balancing the "what" and "where" of image feature extraction. The Channel Attention Module (CAM) focuses on identifying the crucial information about defects, it also helps minimize the impact of channels that contain mostly background information, while the Spatial Attention Module (SAM) pinpoints its precise location. Together, they work in harmony to separate defect features from complex backgrounds and highlight the spatial location of defects in steel images. The architecture of the CBAM model is shown in Figure 8. CAM and SAM modules are linked together in a seamless sequence, bridging the input and output of the CBAM structure with their sophisticated connection. With max and average pooling, the two modules extract rich global and local semantic information. To enhance the feature map from the convolutional network, from the channel dimension, CBAM initially calculates its channel attention template, after which multiplies it by the original feature image to provide the channel a weight recalibration of the original feature image. After that, the extracted feature map is input into the spatial attention module, so that the network learns the significance of different pixel locations in the same channel in an adaptable manner, and finally ultimately acquires the salient features after filtering. Equation (3) and (4) depicts the computation process of each attention map:

$$F' = M_c(F) \otimes F \quad (3)$$

$$F'' = M_s(F') \otimes F' \quad (4)$$

Where $F \in \mathbb{R}^{C \times H \times W}$ indicates as input feature. CBAM applies one-dimensional channel attention map $M_C \in \mathbb{R}^{C \times 1 \times 1}$ and a two-dimensional spatial attention map $M_S \in \mathbb{R}^{1 \times H \times W}$ sequentially to the input $F$. Where $F'$ denotes the new feature map refined by the CAM. Among them, $\otimes$ stands for element-wise multiplication and $F'' \in \mathbb{R}^{C \times H \times W}$ is the final refined output feature. The following describes the details of each attention module.

*1)* Channel Attention Module. Typically, the input image may transfer to a feature matrix through the use of a convolutional layer. The resultant feature matrix's number of channels is the same as the number of kernels, which usually has a value of 1024 or 2048. Because many channel dimensions contain a significant amount of redundancy. To address this, the channel attention model applies a filter to selectively retain only the information most relevant to the desired outcome, thus enhancing the attention on the target. As we can see in the left portion of Figure 6, CBAM utilized both maximum pooling and average pooling to more properly adjust the weight of various channels in the feature map. In short, we can express the whole process by the equation (5):

$$M_C(F) = \sigma(MLP(AvgPool(F)) + MLP(MaxPool)(F)))$$
$$= \sigma\left(W_1\left(W_0(F_{avg}^C)\right) + W_1\left(W_0(F_{max}^C)\right)\right) \quad (5)$$

Where $F$ represents the input's characteristic graph, $F_{max}^C$ denote the max pooling, and $F_{avg}^C$ denote the average pooling operation and then they fed into a shared network consisting of multi-layer perceptron MLP with a single hidden layer. The hidden size of multi-layer perceptron was set to $\mathbb{R}^{C/r}$, to reduce the parameter resources, where $r$ is defined as compression ratio. After applied the shared network, the two output feature vectors are then merged by using element-by-element summation $\oplus$, then merged sum is sent to the Sigmoid activation function $\sigma$ to get the final channel attention force. $W_0 \in \mathbb{R}^{C/r \times C}$, and $W_1 \in \mathbb{R}^{C \times C/r}$ stand for the shared MLP activation operation with ReLu activation function. But in our paper, we have used LeakyReLu activation function [23] instead of ReLu, with its ability to address the "dying ReLU" issue, not only enhances the capacity and accuracy of deep NN but also enables them to achieve more efficient and stable learning.

*2)* Spatial Attention Module: In addition to the preceding channel attention module $M_c(F)$ the spatial attention module $M_s(F)$ completes the process. The spatial attention module typically used to extract the positioning information of essential objects, such as it tells "where" more attention should be paid. The right part of Figure 6 depicts the mechanism of spatial attention module. Both average and max pooling operations are used to compute the spatial attention along the channel axis and then they are concatenated together along the channel dimension. Therefore, SAM can be calculated by the equation (6):

$$M_S(F) = \sigma(f^{7\times7}([AvgPool(F); MaxPool(F)]))$$
$$= \sigma\left(f^{7\times7}([F_{Avg}^S; F_{max}^S])\right) \quad (6)$$

Where $M_S(F) \in \mathbb{R}^{1\times h\times w}$ represent the spatial attention map, the parameter $f^{7\times7}$ indicates the convolution layer with the kernel size of $7 \times 7$, and $\sigma$ denotes the sigmoid activation function. The attention module CBAM is incorporated into our base ResNet50 network, and the integrated ResNet-CBAM is depicted in Figure 8. The convolutional block attention module is applied to the convolution outputs in each block of the ResNet model.

*F. Region of Interest Alignment*

Region of Interest pooling has been an important component in the Faster R-CNN framework for accurately localizing objects within proposed regions of an input image. However, RoI pooling suffers from quantization issues, which can lead to misalignment between the spatial locations of the proposed regions and the corresponding feature map grid cells. RoI Align addresses the issue of misalignment between the input feature map and the region of interest (RoI), which can lead to information loss and reduced precision during the pooling process. By alleviating this misalignment, RoI Align improves object localization accuracy and leads to better performance. The RoI Align process can be summarized as follows:

*A. Input Feature Map*

Given an input image, we first pass it through a convolutional neural network (CNN) to obtain a feature map. This feature map retains spatial information and contains high-level features relevant to object detection.

*B. Region of Interest Proposal*

Object proposals are generated using a region proposal network (RPN). Each proposal is represented by its bounding box coordinates. Subdivision into Grid: For each RoI proposal, RoI Align [24] divides the proposal into a fixed-size grid of evenly spaced cells. This grid is defined independently of the pooling operation and allows us to accurately sample points within the RoI region.

*C. Point Sampling and Bilinear Interpolation*

Unlike RoI Pooling, which quantizes RoI coordinates to discrete bin positions, RoI Align samples points at the original floating-point RoI coordinates. Bilinear interpolation is then applied to extract pixel-level features from the feature map at these points. This interpolation preserves sub-pixel information, resulting in a finer-grained representation. Pooling Process: The features obtained through bilinear interpolation are pooled using an aggregation function such as average or max pooling. The key distinction from RoI Pooling is that RoI Align maintains the continuous nature of the pooled features, preventing information loss due to misalignment.

*D. Output RoI-Aligned Features*

The pooled features from each cell are concatenated to form the final RoI-aligned feature representation. This representation retains detailed spatial information and accurately reflects the content within the RoI. To integrate RoI Align into the Faster R-CNN framework, we replace the conventional RoI Pooling layer in the architecture with the proposed RoI Align module. This seamless substitution allows us to leverage the benefits of RoI Align without requiring extensive modifications to the overall structure of Faster R-CNN.

*E. Focal Loss*

The initial goal of the focal loss function is to address the problem of extreme balance between foreground and background classes during training in object detection scenarios. The starting point of focal loss is the cross-entropy loss function for binary classification, it is defined as:

$$CE(p, y) = CE(pt) = -\log(pt) \quad (7)$$

Where,

$$\begin{cases} p, & if\ y = 1; \\ 1 - p & otherwise \end{cases} \quad (8)$$

$p$ is the prediction probability of the model, and $y$ is ground truth-label. Focal loss is modified from cross-entropy loss by adding a modulating factor $(1 - pt)^\gamma$ to the cross-entropy loss. It is defined as

$$FL(pt) = -(1 - pt)^\gamma \log(pt) \quad (9)$$

In practice, a focal loss function uses an α-balanced variant of focal loss:

$$FL(pt) = -\alpha\ (1 - pt)^\gamma \log(pt) \quad (10)$$

$\gamma$ is a focusing parameter that reshapes the loss function to down-weight easy samples and makes the model focus on hard samples. Hard samples are those samples that produce large errors; a model misclassifies the samples with a high probability.

IV. EXPERIMENTS

In this section, we present the experimental setup conducted on the NEU-DET dataset to evaluate the performance of our proposed approach. The NEU-DET dataset serves as a comprehensive benchmark for our study due to its diversity and complexity, making it an ideal choice for assessing the effectiveness of our methodology. The forthcoming sections will delve into various facets of our experimentation. First, we will provide an in-depth overview of the NEU-DET dataset, detailing its characteristics, data distribution, and annotation intricacies. This comprehensive understanding of the dataset is pivotal for interpreting the subsequent experimental outcomes accurately. Subsequently, we will elucidate the selection of evaluation metrics specifically the COCO [25] evaluation metric that were employed to quantitatively assess the performance of our approach. These metrics offer a comprehensive insight into the methodology's efficacy. Finally, we will expound upon the results obtained from our experiments. These results will be presented in a manner that facilitates an in-depth analysis of the approach's strengths, weaknesses, and overall performance. Moreover, we will provide both qualitative and quantitative assessments to offer a holistic perspective on the outcomes achieved through our proposed methodology. Through this rigorous experimental process, we aim to validate the effectiveness of our approach on the NEU-DET dataset and contribute to a deeper understanding of its potential applications and limitations. The subsequent sections will provide a detailed account of the dataset, evaluation metrics, and experimental outcomes, fostering a comprehensive understanding of the research conducted.

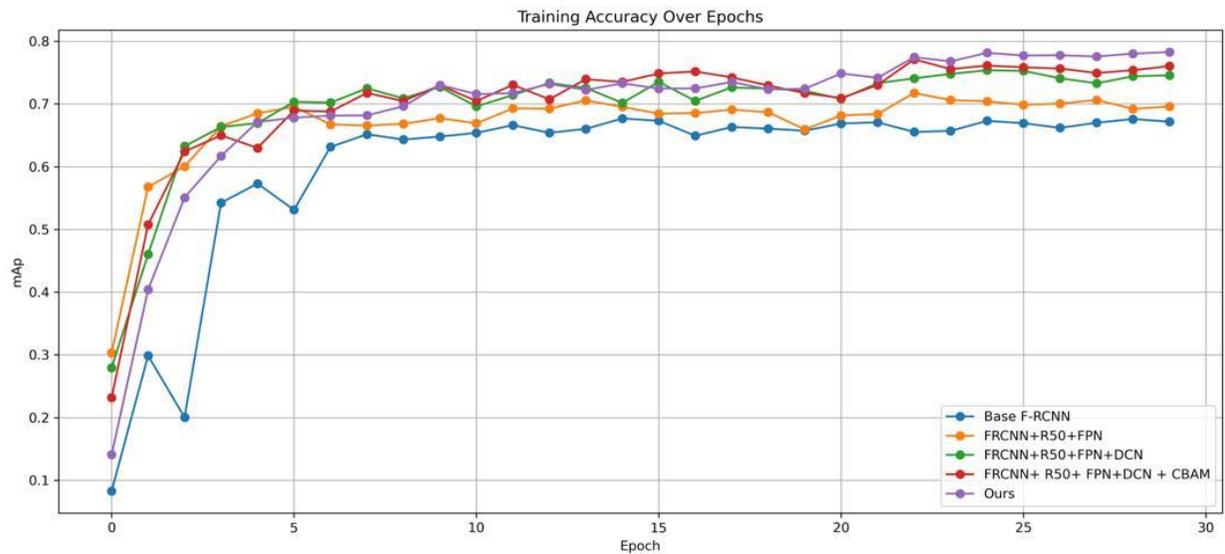

Fig. 10. Training accuracy over epochs of different improved methods.

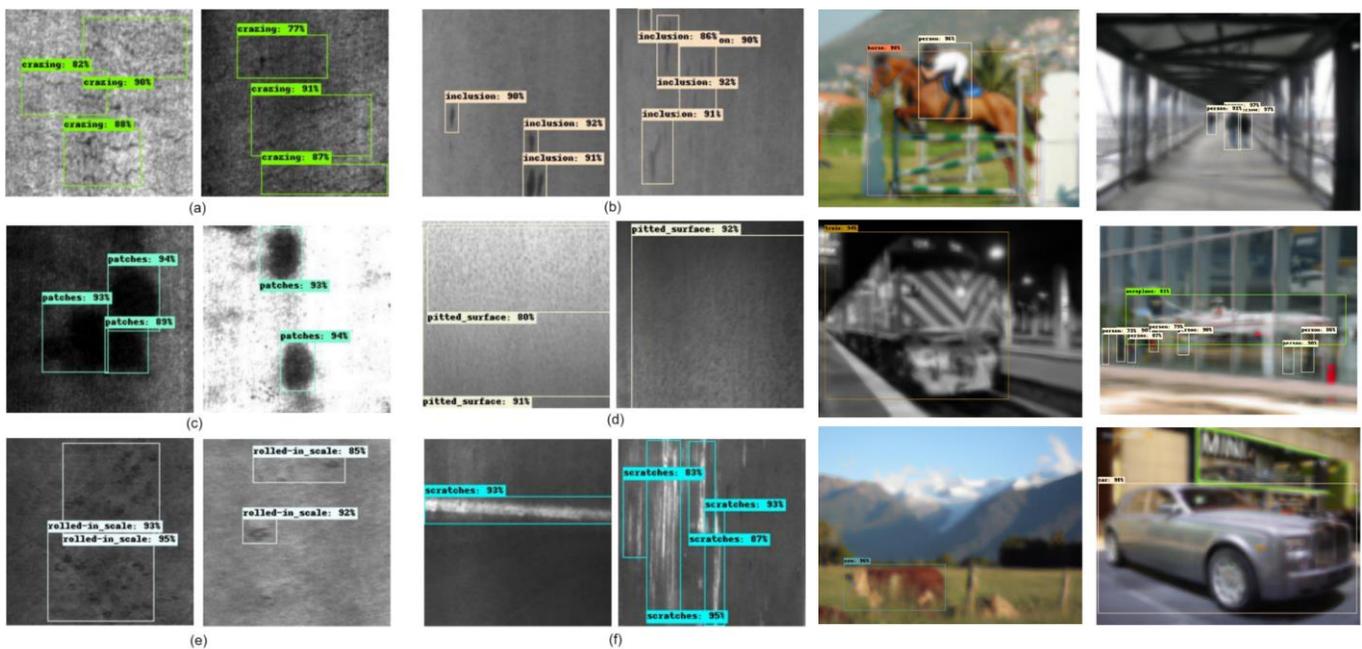

Fig. 11. On the left-hand side: The detection results of six kinds of defects (NEU-DET dataset) based on proposed methods. (a) Crazing, (b) Inclusion, (c) Patches, (d) Pitted_surface, (e) Rolled-in_scale, (f) Scratches. On the right-hand side: Sample predicted

*A. Dataset*

Surface defect images are affected by a variety of factors, such as environment, light, noise, and camera. Furthermore, image quality seriously affects the surface defect detectors.

Table 1: The experimental accuracy of the algorithm on all defects.

| Defects | Crazing | Inclusion | Scratches | Patches | Pitted_surface | Rolled-in_scale |
|---|---|---|---|---|---|---|
| AP | 51.42 | 79.49 | 96.93 | 92.43 | 81.04 | 68.22 |
| **mAP** | | | **78.27%** | | | |

Table 2: Five phases based on different improvement methods.

| Phase | ResNet50+FPN | DCN | CBAM | Focal Loss |
|---|---|---|---|---|
| 1 | × | × | × | × |
| 2 | √ | × | × | × |
| 3 | √ | √ | × | × |
| 4 | √ | √ | √ | × |
| 5 | √ | √ | √ | √ |

However, we focus on how to improve the detection accuracy, not how to get good images. Therefore, we only used the images with good quality. In this work, the NEU-DET dataset is used. NEU-DET consists of six types of surface defects found in hot-rolled steel plates. These defects include patches, rolled-in scale, crazing, inclusion, pitted surface, and scratches. Each type of defect is represented by 300 images, and it is possible for an image to contain multiple defects. The annotations are provided in NEU-DET, which marks the class and location of each defect in an image. Some examples of defect images are shown in Fig 1. In the NEU-DET dataset, there are about 5,000 ground truth boxes. Conventionally, the NEU DET is divided into a training set containing 1440 images and a testing set containing 360 images.

### B. Model Evaluating Metrics

The ResNet-50 with FPN is used as the backbone network of DANet. The NEU-DET dataset is a small dataset which is not enough to support the training of complex neural networks. Therefore, the ResNet-50 is trained on the COCO dataset. All of the surface defect detectors are trained on a Nvidia GeForce RTX 3090, 24G GPU. Stochastic gradient descent (SGD) is used to train with a learning rate of 0.02, momentum of 0.9, batch size is set to 4 and weight decay of 0.0001. We train 30 epochs in an end-to-end manner and decrease the learning rate at the 20th and 28th epochs. The accuracies of all detectors are evaluated by the COCO standard metric average precision ($AP$). The $IoU$ strategy is employed to determine the positive and negative samples from the anchors. The AP is evaluated at a different $IoU$ threshold ($IoU \in [0.5: 0.95]$) over all kinds of surface defects. The precision ($P$) and recall ($R$) are defined as follows:

$$P = \frac{TP}{TP + FP} \quad (11)$$

$$R = \frac{TP}{TP + FN} \quad (12)$$

where $TP$, $FP$, and $FN$ denotes the number of true positive samples, false positive samples, and false negative samples, respectively.

### C. Experimental Results

In this section, we present our baseline object detection model's performance and its relevance for subsequent ablation experiments. The model's defect detection ability is evaluated on real steel surfaces with diverse textures, including six defect types. Results are displayed in Figure 11 using our proposed model, tested on the NEU-DET dataset (Fig. 1. Our baseline model, built on the Faster R-CNN architecture, delivered an initial mean Average Precision (mAP) of 67.60% when evaluated on the NUE-DET dataset. Notably, Table 1 demonstrates the highest average precision for scratches (96.93%) and patches (92.43%), while "crazing" had the lowest (51.42%). Figure 11 (left) illustrates accurate detection and localization of various defects. This baseline serves as a crucial foundation upon which we investigate the incremental impact of algorithmic enhancements. The COCO evaluation revealed the strengths and limitations of the baseline model across diverse defect categories, highlighting the need for improvements to effectively address the intricacies of each defect type. As a preliminary insight, the baseline model's performance varied across defect categories. For instance, "inclusion" detection exhibited higher accuracy compared to "crazing," which proved challenging due to its subtle visual attributes. Similarly, while the model displayed proficiency in detecting "patches" and "scratches," issues were evident in the recognition of "rolled-in_scale" and "pitted_surface," likely due to the complexity of these defect types. These observations underscore the heterogeneous nature of industrial defect detection and lay the foundation for our subsequent ablation experiments. In the subsequent "Ablation Experiments" section, we delve into a meticulous analysis of the effects of algorithmic enhancements on the baseline model's performance. Through a systematic progression, we introduce enhancements such as ResNet50 with FPN, DCN, CBAM, and

Table 3: The experimental results with different improved models are presented, in which average precision with IoU 0.5 is adopted as the experimental index.

| Method | Crazing | Inclusion | Patches | Pitted_surface | Rolled-in_scale | Scratches | mAP |
|---|---|---|---|---|---|---|---|
| Faster R-CNN | 34.10 | 73.40 | 80.93 | 75.29 | 49.93 | 91.95 | 67.60 |
| FRCNN with FPN+R50 | 36.91 | 74.49 | 84.79 | 77.32 | 59.31 | 97.66 | 71.74 |
| FRCNN with FPN+R50+DCN | 46.89 | 75.57 | 88.92 | 80.28 | 63.23 | 97.33 | 75.47 |
| FRCNN with FPN+R50+DCN+CBAM | 52.11 | 79.60 | 87.19 | 83.40 | 63.08 | 97.35 | 77.12 |
| **DANet** (Ours) | 51.42 | 79.49 | 92.43 | 81.04 | 68.22 | 96.93 | **78.27** |

Table 4: Comparisons with other existing models on NEU-DET dataset.

| Method | Crazing | Inclusion | Patches | Pitted_surface | Rolled-in_scale | Scratches | mAP |
|---|---|---|---|---|---|---|---|
| YOLO-v5 | 25.51 | 69.49 | 89.27 | 75.96 | 40.96 | 78.99 | 63.36 |
| RetinaNet | 46.75 | 71.05 | 93.16 | 81.38 | 43.31 | 22.54 | 59.69 |
| CenterNet with Hourglass | 18.38 | 61.07 | 77.17 | 44.58 | 24.25 | 85.58 | 51.83 |
| DETR | 44.64 | **86.19** | 91.67 | 80.90 | 66.54 | 90.99 | 76.82 |
| FCOS | 37.2 | 82.0 | 81.6 | **90.7** | 61.3 | 90.3 | 73.9 |
| TridentNet | 41.0 | 82.9 | **93.4** | 90.3 | 61.6 | 92.5 | 77.0 |
| **DANet** (Ours) | **51.42** | 79.49 | 92.43 | 81.04 | **68.22** | **96.93** | **78.27** |

Focal Loss. By isolating the impact of each enhancement and assessing its effect on various defect categories, we uncover the intricate interplay between algorithmic adjustments and defect attributes. This systematic analysis not only demonstrates the efficacy of each enhancement but also showcases their collective synergy in recognition.

*D. Ablation Study*

The ablation experiments conducted in this study constitute a meticulous exploration of the incremental contributions introduced by various algorithmic enhancements to the foundational Faster R-CNN model. Initially, the base Faster R-CNN architecture yielded a mAP of 67.60% depicted in Table 1 we evaluated our model on COCO evaluation metric. Upon integrating ResNet with FPN in phase-2 shown in Table 3, the detection performance exhibited a discernible advancement, culminating in a mAP of 71.74%. The prowess of FPN in creating a pyramidal feature hierarchy allowed the model to seamlessly capture object details at varying scales, resulting in enhanced object localization. The augmented "Rolled-in_scale" average precision 59.31% and "Scratches" achieved 97.66% after FPN integration stands as a testament to the efficacy of this feature enrichment in capturing the intricate visual attributes of this defect type. Subsequent to the ResNet with FPN enhancement, continuing on our trajectory of enhancements, the introduction of the DCN in phase-3 marked a pivotal milestone in elevating the detection prowess of our model. The subsequent mAP of 75.47% bore testament to the intrinsic adaptability of DCN in capturing deformations inherent to certain defect categories. The "crazing" category's marked improvement from 36.91% to 46.89% underscores the ability adaptive convolutional sampling, where DCN's to dynamically significance of adjust convolutional grids to object contours aids in capturing the intricate crack patterns indicative of "crazing" Paradoxically. The observed decrease in AP for "Scratches" 0.33% might be attributed to the modification in the model's attention patterns introduced by DCN, which might not entirely align with the nuanced features of this particular defect. Following the success of DCN, in phase-4 the integration of the CBAM lent an additional layer of sophistication to our model. The achieved mAP of 77.12% highlighted CBAM's dual focus on channel and spatial attention. This augmentation offered a heightened focus on salient features, contributing to the surge in "crazing" 52.11%, and "inclusion" 79.60% detection accuracy. Paradoxically, the AP of "Patches" and "Rolled-in_scale" exhibited a decrease 1.73% and 0.15%, respectively, likely due to the fact that the attention mechanisms might not align well with the distinctive textural attributes of this defect type. The final augmentation, Focal Loss, played a pivotal role in refining the model's localization prowess, leading to the highest attained mAP. By addressing the issue of class imbalance, the introduction of Focal Loss culminated in an overall mAP improvement of 1.15%, reaching an impressive 78.27%. Noteworthy is the substantial enhancement in "Patches" detection, with a 5.24% increase in AP, resulting in a final value of 92.43%. Similarly, "Rolled-in_scale" detection exhibited a commendable improvement of 5.14%. However, it's important to note that these improvements were not uniform across all defect types. While the 'patches' and 'rolled-in_scale' defect categories experienced substantial enhancements, some other defect types saw a slight decline in their AP scores. This observation suggests that the adjustments made to cater to certain defect types might have inadvertently affected the model's performance on others. The comprehensive results of these ablation experiments provide not only insights into the unique strengths of each enhancement but also highlight the intricate interplay that ultimately contributes to a robust and versatile object detection model for the nuanced landscape of industrial defect recognition. Furthermore, we assessed the performance of our DANet model using a subset of 2000 images from the Pascal VOC dataset, owing to resource limitations. Our DANet model outperformed the base Faster R-CNN model significantly, achieving a mAP of 0.574, compared to the base Faster R-CNN's mAP of 0.372. This result highlights the effectiveness of DANet in improving small and complex object detection accuracy on the dataset.

*E. Comparision*

In comparison to established models such as YOLOv5, RetinaNet, DETR, TridentNet[26], FCOS[27], and CenterNet, our proposed model, referred to as DANet, stands out as a remarkable advancement in object detection performance. We evaluate these models using the mean Average Precision (mAP) metric, a widely accepted measure of accuracy in object detection tasks. DANet achieved an impressive mAP value of 78.27, surpassing all other models under consideration. YOLOv5, a well-known model in the field, demonstrated a respectable performance with a mAP score of 63.36 after train 100 epochs. While YOLOv5 exhibited competitive results, it is noteworthy that our DANet model outperformed it by a substantial margin depicted in Table 4, emphasizing DANet's efficacy in detecting objects within various contexts. DETR, another prominent model in object detection, achieved a mAP score of 76.82[28]. However, a distinct observation emerges when examining specific defect types such as "inclusion". In this defect categories, DETR exhibited significantly higher Average Precision (AP) scores compared to our proposed DANet. This discrepancy suggests that DETR might excel in detecting certain defect characteristics that might be challenging for DANet to discern accurately. CenterNet, on the other hand, displayed the lowest overall performance among the models evaluated, attaining a mAP value of 51.83. This outcome further underscores the competitiveness of our DANet model, as it outshines CenterNet with a substantial lead, demonstrating its robustness in handling object detection tasks. In conclusion, our proposed DANet model emerges as the frontrunner in this comparative analysis, achieving the highest mAP value of 78.27 when pitted against YOLOv5, TridentNet[29], FCOS[29]. While FCOS showed prowess in certain defect categories, DANet's overall performance superiority reinforces its potential as a groundbreaking solution in object detection applications, showcasing its capacity to address a wide array of detection challenges with remarkable accuracy.

## V. Conclusion

In conclusion, this research aimed to enhance the accuracy of defect detection using the Faster R-CNN model along with additional improvement algorithms. The results of the experiments demonstrated promising progress in identifying defects within the dataset. However, it is evident that certain defects posed challenges in terms of accurate detection, indicating the complexity and variability inherent in real-world defect scenarios. Despite the success achieved, there remains scope for further refinement. Looking ahead, the field of defect detection stands to benefit from the integration of more advanced algorithms, such as transformers, which have shown remarkable capabilities in various computer vision tasks. By harnessing the potential of these cutting-edge techniques, we aim to enhance the accuracy and robustness of defect detection systems, thus addressing the persisting challenges identified in this study. Moreover, expanding the scope of experimentation to encompass a wider range of defect datasets will provide a more comprehensive evaluation of the proposed methods and their adaptability across different scenarios. As the field of computer vision and machine learning continues to evolve, it is crucial to recognize that defect detection is an ongoing challenge that demands constant innovation. While this research has made noteworthy strides, there is an exciting journey ahead involving the exploration of novel methodologies, the refinement of existing techniques, and the accumulation of a richer dataset. Ultimately, by combining advanced algorithms with an increasingly diverse array of defect data, we can aspire to develop defect detection systems that are more accurate, reliable, and adaptable to a wide array of real-world applications.

Acknowledgment

We extend our heartfelt gratitude to Professor Chunbiao Li for his invaluable feedback and generous provision of GPU resources, which greatly enhanced the quality and depth of our experiments.